%% file: main.tex
\documentclass[letterpaper]{article} % DO NOT CHANGE THIS
\usepackage{aaai25}  % DO NOT CHANGE THIS
\usepackage{times}  % DO NOT CHANGE THIS
\usepackage{helvet}  % DO NOT CHANGE THIS
\usepackage{courier}  % DO NOT CHANGE THIS
\usepackage[hyphens]{url}  % DO NOT CHANGE THIS
\usepackage{graphicx} % DO NOT CHANGE THIS
\usepackage{natbib}  % DO NOT CHANGE THIS AND DO NOT ADD ANY OPTIONS TO IT
\usepackage{caption} % DO NOT CHANGE THIS AND DO NOT ADD ANY OPTIONS TO IT
\usepackage{algorithm}
\usepackage{algorithmic}

\usepackage{newfloat}
\usepackage{listings}
\DeclareCaptionStyle{ruled}{labelfont=normalfont,labelsep=colon,strut=off} % DO NOT CHANGE THIS
\floatstyle{ruled}
\newfloat{listing}{tb}{lst}{}
\floatname{listing}{Listing}
\nocopyright % -- Your paper will not be published if you use this command
\usepackage{epsfig}
\usepackage{amsmath}
\usepackage{amssymb}

\usepackage{flexisym}
\usepackage{multirow}
\usepackage{subcaption}

\usepackage{tikz}
\usepackage{comment}

\usepackage{color, colortbl}

\usepackage{booktabs}

\usepackage[accsupp]{axessibility}  % Improves PDF readability for those with disabilities.

\DeclareMathOperator*{\argmin}{arg\,min}

\newcommand{\Skip}[1]
{
}

\makeatletter
\def\@fnsymbol#1{\ensuremath{\ifcase#1\or \dagger\or \ddagger\or
\mathsection\or \mathparagraph\or \|\or **\or \dagger\dagger
\or \ddagger\ddagger \else\@ctrerr\fi}}
\makeatother

\title{Long-term Pre-traning for Temporal Action Detection with Transformers}
\author{
    Jihwan Kim\quad\quad\quad Miso Lee\quad\quad\quad Jae-Pil Heo\thanks{Corresponding author}
}
\affiliations{
    {\Large
    Sungkyunkwan University}\\
    \vspace{3.0pt}
    {\tt\small \{damien,\;dlalth557,\;jaepilheo\}@skku.edu}
}

\begin{document}

\maketitle

\input{sec/0_abstract}


\input{sec/1_introduction}
\input{sec/2_related_work}
\input{sec/3_our_approach}
\input{sec/4_experiments}
\input{sec/5_conclusion}

\bibliography{aaai25}

\newpage

\input{sec/7_supplementary}

\end{document}

%% file: sec/0_abstract.tex
\begin{abstract}
Temporal action detection (TAD) is challenging, yet fundamental for real-world video applications.
Recently, DETR-based models for TAD have been prevailing thanks to their unique benefits.
However, transformers demand a huge dataset, and unfortunately data scarcity in TAD causes a severe degeneration.
In this paper, we identify two crucial problems from data scarcity: attention collapse and imbalanced performance.
To this end, we propose a new pre-training strategy, Long-Term Pre-training (LTP), tailored for transformers.
LTP has two main components: 1) class-wise synthesis, 2) long-term pretext tasks.
Firstly, we synthesize long-form video features by merging video snippets of a target class and non-target classes.
They are analogous to untrimmed data used in TAD, despite being created from trimmed data.
In addition, we devise two types of long-term pretext tasks to learn long-term dependency.
They impose long-term conditions such as finding second-to-fourth or short-duration actions.
Our extensive experiments show state-of-the-art performances in DETR-based methods on ActivityNet-v1.3 and THUMOS14 by a large margin.
Moreover, we demonstrate that LTP significantly relieves the data scarcity issues in TAD.
\end{abstract}

%% file: sec/1_introduction.tex
\vspace{-10pt}
\section{Introduction}
Understanding videos has become fundamental for real-world video applications with the proliferation of videos being shared across online platforms. 
Initially, researchers concentrated on classifying actions in videos by using trimmed clips.
However, this became expensive, leading to the development of Temporal Action Detection (TAD). 
TAD not only classifies actions correctly but also identifies the time boundaries in untrimmed videos.

End-to-end object DEtection with TRansformers (DETR)~\cite{carion2020detr} has been prevailing in the literature of detection.
DETR views detection as a set prediction for end-to-end detection without any human prior like non-maximum suppression or pre-defined anchors.

\begin{figure}[t]
\centering
\includegraphics[width=8.35cm]{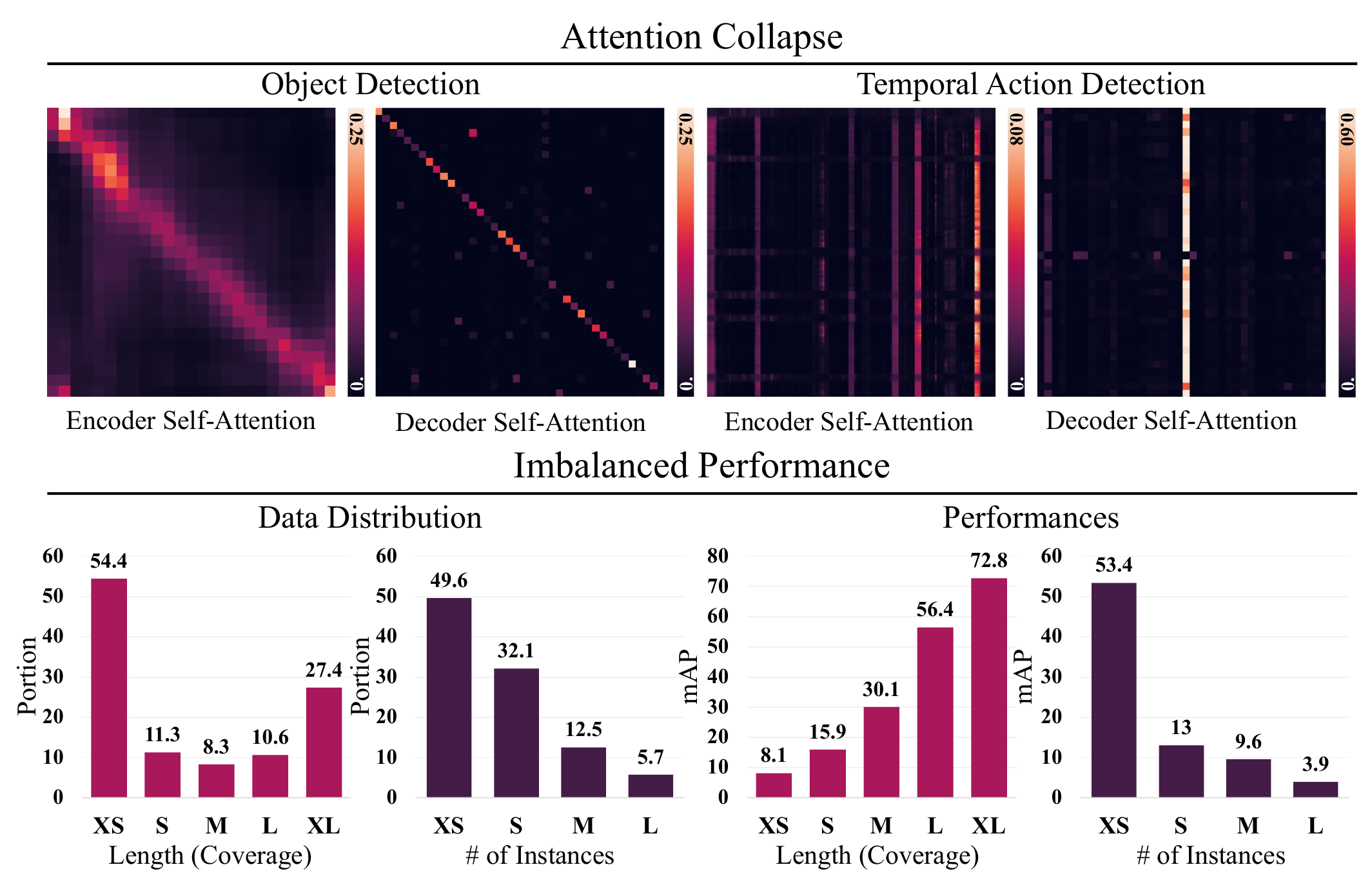}
\vspace{-15pt}
\caption{\textbf{Problems from data scarcity.}
There are two main problems caused by the data scarcity in DETR for TAD: attention collapse and imbalanced performance.
The first row shows the collapsed self-attention maps from encoder and decoder of DAB-DETR.
The second depicts imbalanced performances in terms of action lengths (in Coverage) and the number of instances on ActivityNet-v1.3 from DETAD.
}
\label{fig:introduction}
\vspace{-10pt}
\end{figure}

Introduction of DETR to TAD is natural for next research steps as DETR inherits strengths of transformer~\cite{vaswani2017attention}, bringing two valuable characteristics for TAD.
First, it does not depend on pre-defined anchors.
This merit is crucial for TAD since actions in videos have more diverse time lengths than object sizes.
Second, it can directly learn global relation without stacking multiple layers.
It can relieve the problem from long-term dependency for TAD.

Despite the great benefits, transformers demand a large-scale dataset with sufficient inter- and intra-class diversities to learn dense relations without an inductive bias.
Especially for DETR, the bipartite matching further strengthens the demand because of the label inefficiency from one-to-one matching.
However, the current TAD benchmarks do not meet such a demand due to expensive labeling cost.

Consequently, directly employing DETR to TAD causes a severe degeneration for transformers.
We empirically identify two primary problems from data scarcity: 1) attention collapse, 2) imbalanced performance, as depicted in Fig.~\ref{fig:introduction}.
Firstly, the collapse of self-attention in transformer-based methods is the phenomenon to skip the self-attention layers since the model cannot learn expressive relations between video features and action queries from data~\cite{dong2021rank_collapse}.
As in the upper of Fig.~\ref{fig:introduction} unlike DETR for object detection, all queries of DETR for TAD attend a few key elements in which the gradients for them are vanishing~\cite{noci2022signal}, indicating that the self-attention quits learning.
The collapse problem can be caused by the two main aspects: model architecture~\cite{dong2021rank_collapse, zhai2023stabilizing} and data diversity~\cite{tang2021augmented, trockman2023mimetic}.
In this paper, we focus on the data perspective to cope with the collapse problem.
Secondly, the model exhibits severe imbalanced performances in terms of action lengths and the number of action instances within a video.
DETR does not use multi-scale division for label assignment, in which all queries are responsible for all temporal scales of actions.
When this design faces insufficient data, it can be easier to fall in over-fitting to a major scale (usually to long actions).

To settle down the issue, we simply suggest that DETR be pre-trained with large-scale balanced data.
Here, our focus is on `the detector (i.e., DETR)', not on the feature extractor.
Although there are many attempts to pre-train the feature extractor, there is no consideration for the detector.
The pre-training for the detector should be oriented towards the temporal detection as it is designed for localization.

To this end, we develop a new pre-training strategy tailored for transformers of DETR, named Long-Term Pre-training (LTP).
LTP has two main components: 1) class-wise synthesis, 2) long-term pretext tasks.
Firstly, we synthesize long-form video features for pre-training from a large-scale video-classification dataset with trimmed clips to provide the model with sufficient and balanced data diversity.
We build pre-training features based on action categories to minimize task discrepancy between the pretext task in pre-training and the downstream task (TAD) in fine-tuning.
Although our way of synthesizing is closely related to previous pre-training methods, the key difference is in the class-wise pretext task, for which the model should detect all actions of the target class, instead of identifying a single action instance.
Secondly, we further introduce two long-term pretext tasks, ordinal and scale conditions, to teach the model long-term dependency.
For example, the model is required to identify exclusively “first-to-second” actions (ordinal) or “short-duration” actions (scale) among instances of the target category in synthesized videos during pre-training.
These tasks impose specialized conditions on the basic pretext task, to further encourage DETR to learn long-term dependency and diverse relations in attention.

Our extensive experiments demonstrate that our pre-training remarkably alleviates the data scarcity problem.
As a result, DETR models with LTP outperform the state-of-the-art DETR-based methods on two TAD benchmarks, ActivityNet-v1.3 and THUMOS14, by a large margin.

To sum up, our main contributions are as follows:
\begin{itemize}
	\item To our knowledge, this is the first attempt to highlight the importance of pre-training DETR for TAD in mitigating the issues arising from limited data.
	\item We propose a novel pre-training strategy, Long-Term Pre-training (LTP) for TAD with transformers. 
    Class-wise synthesis and long-term pretext tasks provide diverse and balanced data and capability to learn long-term dependency while minimizing task discrepancy.
	\item Our extensive experiments demonstrate that LTP significantly relieves the data scarcity issues with a new SoTA performance on ActivityNet-v1.3 and THUMOS14.
\end{itemize}

%% file: sec/2_related_work.tex
\section{Related Work}
\subsection{Temporal Action Detection}
Temporal action detection (TAD) aims to find time intervals of actions of interest in untrimmed videos as well as classifying the instance.
Pioneering methods~\cite{yeung2016frame-glimpses, shou2016scnn, buch2017sst} have been made great advance in TAD during the past decade.
As two-stage mechanism had been successful in object detection, numerous methods in TAD deployed multi-stage framework~\cite{gao2017turn, zhao2017ssn, xu2017rc3d, kim2019coarsefine}.

As the following methods, point-wise learning has been widely adopted to generate more flexible proposals without pre-defined time windows.
SSN~\cite{zhao2017ssn} and TCN~\cite{dai2017tcn} expanded temporal context around a generated proposal to improve ranking performance.
BSN~\cite{lin2018bsn} and BMN~\cite{lin2019bmn} grouped possible start-end pairs to build action proposals, then scored them for final localization predictions.
BSN++~\cite{su2021bsn++} tackled the imbalance problem over temporal scales based on BSN.
Recently, ActionFormer~\cite{zhang2022actionformer} deployed transformer-based encoder as multi-scale backbone network, and BRN~\cite{kim2024boundary} resolved the issue of multi-scale features for TAD.

\subsection{Pre-training for Detection}
End-to-end object DEtection with TRansformers (DETR)~\cite{carion2020detr} firstly viewed object detection as a direct set prediction problem, and removed the need of human heuristics like non-maximum-suppression (NMS).
However, transformers of DETR typically need 10 times more training iterations than the conventional detectors due to bipartite matching and dense attention.

\begin{figure}[t]
\centering
\includegraphics[width=8.35cm]{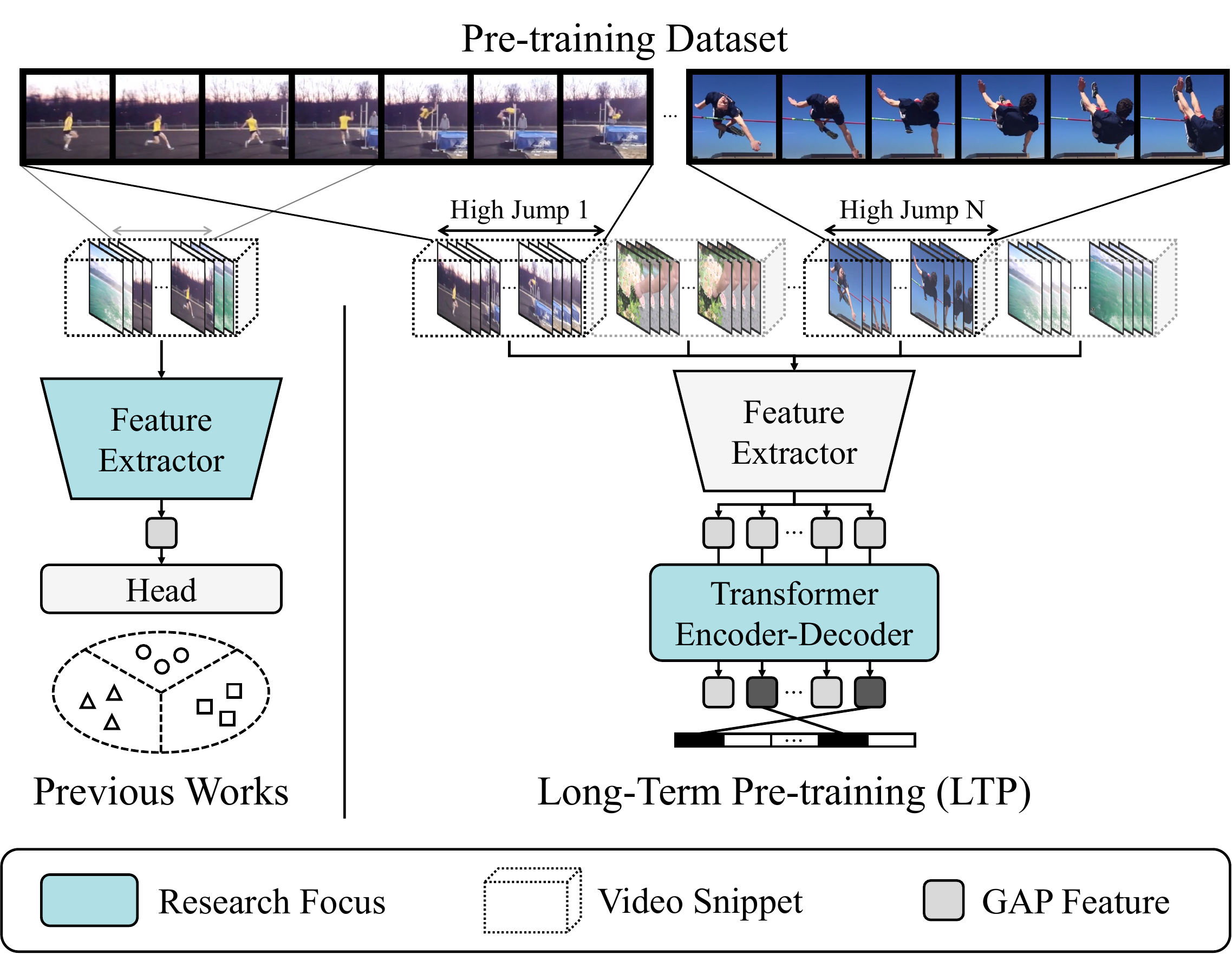}
\caption{\textbf{Differences b/w previous and our pre-training.}
Previous pre-training focused on the feature extractor.
However, there has been no research conducted on pre-training DETR for TAD despite the issues on the data scarcity.
The pretext tasks to train feature extractor and detector should be different.
LTP is designed for class-wise localization from long-form videos, just like the downstream task.
}
\label{fig:difference}
\end{figure}

\begin{figure*}[t]
\centering
\includegraphics[width=17.75cm]{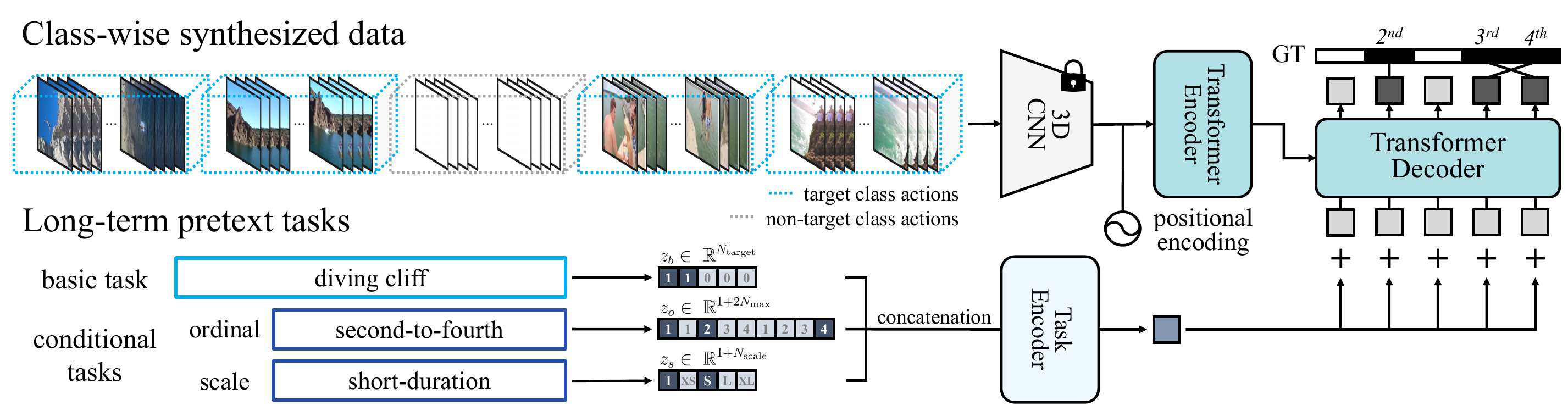}
\caption{\textbf{Overall procedure of Long-Term Pre-trainig (LTP).}
LTP has two main components: 1) class-wise synthesis, 2) long-term pretext tasks.
Class-wise synthesis aims to minimize the task discrepancy by building training features to localize based on categories.
Moreover, the conditional tasks aim to learn the long-term dependency by ordinal or scale conditions.
}
\label{fig:architecture}
\end{figure*}

To address the issue, there have been research efforts~\cite{dai2021up-detr, bar2022detreg} for pre-training DETR.
UP-DETR~\cite{dai2021up-detr} is the first attempt for pre-training DETR to overcome data scarcity.
UP-DETR introduced an instance-wise pre-training strategy, which finds the instance specified by the pooled features of a random region in the image.
Also, DETReg~\cite{bar2022detreg} pointed out that the randomness of UP-DETR reduces the benefits of pre-training and to resolve this problem they devised object-centric selection in an unsupervised manner.

In the literature of TAD, there have been also pre-training approaches~\cite{alwassel2021tsp, xu2021bsp, xu2021lofi, zhang2022pal, kang2023sola} but all for the feature extractor.
TSP~\cite{alwassel2021tsp} and BSP~\cite{xu2021bsp} have proposed boundary-sensitive pre-training strategies for the feature extractor.
Also, LoFi~\cite{xu2021lofi} devised fine-tuning strategies for the feature extractor with fewer clips and small images.
On the other hand, PAL~\cite{zhang2022pal} and SoLa~\cite{kang2023sola} have introduced the methods for the task-specific temporal encoder for the feature extractor.

Recently, DETR has been introduced in TAD~\cite{tan2021relaxed, liu2021tadtr, shi2022react, kim2023self} as it prevails in object detection.
However, there are no research attempts of pre-training DETR for TAD while the performance of DETR for TAD has been far behind conventional TAD methods due to the severe data scarcity.
As for the feature extractor, the goal is to teach frame-wisely fine-grained temporal sensitivity with short videos ($\sim$2.6 seconds) as illustrated in Fig.~\ref{fig:difference}.
On the other hand, the pre-training for the detector should be oriented towards the temporal detection as it is designed for localization.
Unlike previous pre-training methods of TAD, our pre-training method is specialized for pre-training of DETR for TAD.

%% file: sec/3_our_approach.tex
\section{Our Approach}
In this section, we elaborate the details of our pre-training framework, Long-Term Pre-training (LTP).
As illustrated in Fig.~\ref{fig:architecture}, LTP has two main components: 1) class-wise synthesis, 2) long-term pretext tasks.
In class-wise synthesis, we construct features to form a localization problem by integrating target and non-target actions.
Additionally, we introduce two kinds of conditions: ordinal and scale tasks.

\subsection{Preliminary}
% DETR
\noindent\textbf{DETR.} DETR~\cite{carion2020detr} is an architecture designed for end-to-end detection, comprising feature extractor, transformer encoder, and transformer decoder.
The feature extractor captures general features from images, while the transformer encoder and decoder focus on encoding more specific features and decoding object information from features.
The core of the transformer is the attention mechanism, which is defined formally as follows:
\begin{equation}
    \begin{aligned}
    \text{Attention}(Q, K, V)=AV, \\
    A=\text{softmax}(\frac{QK^{\top}}{\sqrt{d_{k}}}),
    \label{eq:attention}
    \end{aligned}
\end{equation}
where $Q$, $K$, $V$, $d_k$ are query, key, value and dimension of the key, respectively.
When we feed the same input into query, key, and value, it is commonly referred as self-attention mechanism.
The transformer encoder in DETR consists of multiple layers of self-attention and feed-forward networks, refining the CNN features for the task.

On the other hand, the cross-attention mechanism employs the same key and value but with a different query.
This enables the query to be represented by the input given as key and value.
The transformer decoder in DETR includes several layers of self-attention, cross-attention, and feed-forward networks.
Learnable embedding vectors are fed into self-attention networks as input and cross-attention networks as query, so called as decoder queries.
For the cross-attention networks, encoder outputs are used as key and value.
This allows the decoder to extract unique localization information from the encoder outputs for each query.
Finally, DETR predicts the classes and the locations using the decoder outputs and bipartite matching.

\vspace{3pt}
\noindent\textbf{DETR for TAD.}
In TAD, we deploy a 3D CNN as the feature extractor, which is pre-trained on Kinetics~\cite{kay2017kinetics}.
Note that the 3D CNN is fixed while training the transformer encoder and decoder of DETR.
To extract the video features, each video is fed into the 3D CNN followed by global-average pooling for spatial dimensions so that only the temporal dimension remains.
When it comes to architectures of DETR, we deploy two representative architectures, DAB-DETR~\cite{shilong2022dab_detr} and deformable-DETR~\cite{xizhou2021deformable_detr} in this paper. 
Decoder queries stand for action instances, called action queries.
Therefore, the decoder receives the refined video features from the encoder and relates them with action queries.
Finally, the output of the decoder passes through classification and regression heads, then final detection results are produced.

\subsection{Class-wise Synthesis}
\noindent\textbf{Motivation.} 
In order to provide sufficient and balanced data especially for relieving imbalanced performances over scales, we utilize large-scale action recognition dataset like Kinetics~\cite{kay2017kinetics}.
Also, long-form video features are vital to build a localization pretext task for transformer in DETR.
Here, we aim to minimize the task discrepancy between pre-training and fine-tuning (TAD) while making large-scale balanced data via class-wise synthesis.

\vspace{3pt}
\noindent\textbf{Synthesis.} 
As with earlier pre-training approaches~\cite{alwassel2021tsp, xu2021bsp}, we must create video features suitable for localization since action recognition videos lack background regions.

To achieve this, we bring in class-wise synthesis to reduce task differences.
We start by picking a single target action category from the Kinetics classes. 
Then we randomly select videos from other classes to make a background template, which holds non-target actions.
Once the template is set, we randomly choose videos from the target class.
By placing these chosen target category videos into the background template, we create synthesized video features for the pretext task.

Formally, let us denote the set of categories in Kinetics as $K=\{1,...,N_K\}$.
Each video in Kinetics only contains a single action category among $K$.
Also, we have a pre-trained feature extractor $F(\cdot)$, which fixed across both pre-training and fine-tuning phases.
Initially, a random selection is made from $K$ to determine the target category $k^\ast$.
In order to construct a background template $B$ with only non-target actions, we randomly sample $N_{bg}$ videos of other classes except for the target category as below:
\begin{equation}
B = [F(v_1), F(v_2), ..., F(v_{N_{bg}})], \quad k_{v_i} \neq k^\ast,
\end{equation}
where $F(v_i)$ is temporal features of the video $v_i$ after global-average pooling, $[\cdot]$ means concatenation along time axis, and $k_{v_i}$ is the category of the video $v_i$.

After forming the template, we also randomly sample $N_t$ videos of the target class as below:
\begin{equation}
T = \{F(v_1), F(v_2), ..., F(v_{N_t})\}, \quad k_{v_i} = k^\ast,
\end{equation}
where $\{\cdot\}$ indicates a set of features.

With the sampled target features $T$, we firstly temporally crop each video feature into a random size.
Then we insert the cropped features of the sampled target videos into the background template $B$ at random temporal positions.
We utilize all $N_t$ target features and allow overlapping.

\subsection{Long-term Pretext Tasks}
\noindent\textbf{Motivation.} 
Alongside the basic task to localize actions of the target class, we introduce two types of long-term pretext tasks: ordinal task, and scale task.
These tasks are designed to reduce the degree of the attention collapse problem.
The tasks encourage the model to learn long-term dependency, and prevent it from attending to a few of temporal features.

\vspace{3pt}
\noindent\textbf{Basic Task.} 
To appropriately guide the model in our pre-training, we need to define the desired objective by specifying the target class.
This target category is represented through one-hot encoding, denoted as the one-hot vector $z_{b}$ for indicating the target class for the basic pretext task.
The dimension of $z_{b}$ is in $\mathbb{R}^{N_{\text{target}}}$, where $N_{\text{target}}$ corresponds to the total number of action classes in $K$.

\vspace{3pt}
\noindent\textbf{Conditional Tasks.} 
In addition to the basic task, we introduce two types of pretext tasks: ordinal and scale pretext tasks.
Firstly, the ordinal task involves identifying actions falling within specified ordinal number ranges.
For instance, the model can be trained to identify the third-to-fifth actions of the target class.
The ordinal task prompts the model to understand the relative positioning of actions within the video sequence.
This requires the model to consider not only individual actions but also their sequence, encouraging the learning that extend beyond neighboring actions.

Specifically, the start ordinal number and range length are randomly selected based on the current number of target actions within the synthesized features.
Note that ordinals are assigned exclusively to target class actions.
Through the ordinal task, the DETR decoder's action queries are prompted to consider other queries for locating the ordinal position.
This fosters the model to learn the global relationships within the video.

Secondly, the scale pretext task focuses on localizing actions corresponding to defined temporal scales.
Understanding scale involves how relatively long an action lasts in the video sequence.
This comprehension prompts the model to understand the speed of actions compared to the neighboring actions and the global context.

Concretely, we define four scales: extra-short (XS), short (S), long (L), and extra-long (XL).
These scales are determined by the action's duration-to-total-video-length ratio ($r$).
XS actions are characterized by $r < 0.25$, S by $0.25 \leq r < 0.50$, L by $0.50 \leq r < 0.75$, and XL by $0.75 \leq r$.
To designate the scale, we randomly select one target action from the synthesized features and assign the task with the corresponding scale.

Similar to the basic task, we introduce conditions through one-hot encoding.
Formally, let us denote the one hot vectors for the ordinal and scale tasks as $z_o$, and $z_s$, respectively.
In the scale pretext task, the one-hot vector $z_s$ has a length of $5$.
The initial channel of $z_s$ indicates whether to apply the scale condition or not, while the remaining four channels specify the target scale among the four available scales.
For instance, $[1, 0, 1, 0, 0]$ describes the intent to identify short-duration actions within the target class.
Here, we only assign a single scale for the condition.

Moreover, when the maximum number of target action instances within a video in pre-training is denoted as $N_{\text{max}}$, the one-hot vector $z_o$ encompasses $2N_{\text{max}} + 1$ channels.
Likewise, the first channel serves as an indicator for conditioning.
The subsequent $N_{\text{max}}$ channels following the indicator mean the first ordinal number for the range, while the final $N_{\text{max}}$ channels designate the second ordinal number.
For example, with $N_{\text{max}}$ set to 4, the one-hot vector for the ordinal task, seeking second-to-fourth actions within the target class, would be $[1, 0, 1, 0, 0, 0, 0, 0, 1]$.

To establish conditional tasks alongside the basic pretext, we concatenate all one-hot vectors for both basic and conditional pretext tasks along the channel axis.
Whether to condition or not is assigned with a probability $p_{\text{cond}}$ in uniform distribution.
We also employ a task encoder $E(\cdot)$ consisting of three fully-connected layers with ReLU activations.
This task encoder $E(\cdot)$ aligns the dimensionality of the one-hot vector to match the action queries of DETR.
These projected vectors are then added to DETR decoder's action queries.
Formally, let the decoder's action queries be $q$.
The conditioned queries, denoted as $q'$, are defined as follows:
\begin{equation}
q' = q \oplus E([z_{b}, z_{o}, z_{s}]),
\end{equation}
where $\oplus$ is element-wise summation, and $[\cdot]$ is concatenation along the channel axis.

\subsection{Objectives}
\noindent\textbf{DETR.}
We follow the objectives in DETR~\cite{carion2020detr}. 
Let us denote the ground-truths, and the $M$ predictions as $y$, $\hat{y}={\hat{y_i}_{i=1}^{M}}$, respectively.
For the bipartite matching between the ground-truth and prediction sets,
we define the optimal matching $\hat{j}$ with the minimal cost and search from all possible permutation of $\{1,...,M\}$ denoted by $J_M$ as follows:
\begin{equation}
    \begin{aligned}
    \hat{j} = \argmin_{j \in J_M} \sum_{i}^{M}\mathcal{L}_{\text{match}}(y_i, \hat{y}_{j(i)}),
    \label{eq:bipartite_matching}
    \end{aligned}
\end{equation}
where $L_{\text{match}(y_i, \hat{y}_{j(i)})}$ is a pair-wise matching cost between $y_i$ and the prediction with the index from $j(i)$, which outputs the index $i$ from the permutation $j$.

Next, let us denote each ground-truth action as $y_i = (c_i, t_i)$, where $c_i$ is the target class label with the background one $\emptyset$, and $t_i$ is the time intervals.
For the prediction with the index $j_{(i)}$, we define the probability of the class $c_i$ as $\hat{p}_{j(i)}(c_i)$ and the predicted time intervals as $\hat{t}_{j(i)}$.
Then we formulate the main objective as following:
\begingroup
\small{
\begin{equation}
    \begin{aligned}
    \mathcal{L}_{\text{DETR}}(y, \hat{y}) = \sum_{i=1}^{M}[-\log{\hat{p}_{\hat{j}(i)}(c_i)} + \mathbb{1}_{c_i \neq \emptyset}\mathcal{L}_{\text{reg}}(t_i, \hat{t}_{\hat{j}(i)})],
    \label{eq:detr_loss}
    \end{aligned}
\end{equation}
}
\endgroup
where $\hat{j}$ is the optimal assignment from Eq.~\ref{eq:bipartite_matching}, and $\mathcal{L}_{\text{reg}}(t_i, \hat{t}_{j(i)})$ is the regression loss between the ground-truth $t_i$ and the prediction $\hat{t}$ with the index $j(i)$.

\vspace{3pt}
\noindent\textbf{Pre-training.}
In LTP, class-wise synthesis assigns the target category to the model, so that the DETR model to perform binary classification.
The class-agnostic approach enhances model generality as in class-agnostic object detection~\cite{maaz2022mavl}.
To achieve this, we adopt the identical bipartite matching objective as DETR, as depicted in Eq.~\ref{eq:detr_loss}, with binary classification labels.
This configuration enables the model to address the localization task in a class-agnostic manner.
Apart from classification, the remaining components of the objective remain consistent with $\mathcal{L}_{\text{DETR}}$.

In our conditional tasks, we delineate actions to locate based on ordinal and scale criteria.
These conditions serve to narrow down the scope, aiming to focus on specific target actions based on the criteria.
For instance, specifying a search for first-to-fourth actions solely involves the target actions, excluding non-target ones from the ordinal count.
Similarly, in cases where the scale condition pertains to identifying short-duration actions, the model must pinpoint short actions within the target class, not short instances from other classes.
Consequently, the basic task consistently serves as the objective, while conditional tasks are randomly enabled or disabled with a probability of $p_{\text{cond}}$.

%% file: sec/4_experiments.tex
\section{Experiments}
\begingroup
\setlength{\tabcolsep}{4.95pt} % Default value: 6pt
\renewcommand{\arraystretch}{1.0} % Default value: 1
\begin{table*}[t]
\centering
\begin{tabular}{l||ccccc|c||ccc|c}
    \hline\hline
    \multirow{2}{*}{Method} &
    \multicolumn{6}{c||}{THUMOS14} & \multicolumn{4}{c}{ActivityNet-v1.3} \\
    \cline{2-11}
    & $0.3$ & $0.4$ & $0.5$ & $0.6$ & $0.7$ & Avg. & $0.5$ & $0.75$ & $0.95$ & Avg. \\ 
    \hline\hline
    \rowcolor{gray!25}\multicolumn{11}{l}{\textit{\textbf{DETR-based Methods}}} \\
    \hline\hline
    RTD-Net~\cite{tan2021relaxed} & $68.3$ & $62.3$ & $51.9$ & $38.8$ & $23.7$ & $49.0$ & $47.21$ & $30.68$ & $8.61$ & $30.83$ \\
    ReAct~\cite{shi2022react} & $69.2$ & $65.0$ & $57.1$ & $47.8$ & $35.6$ & $55.0$ & $49.60$ & $33.00$ & $8.60$ & $32.60$ \\
    \hline
    DAB-DETR~\cite{shilong2022dab_detr} & $70.5$ & $64.3$ & $53.9$ & $39.3$ & $23.8$ & $50.3$ & $50.62$ & $32.59$ & $7.43$ & $32.40$ \\
    DAB-DETR + LTP (Ours) & $75.2$ & $69.5$ & $60.4$ & $48.4$ & $32.8$ & $57.3$ & $52.71$ & $35.60$ & $9.08$ & $34.92$ \\
    \hline
    Self-DETR~\cite{kim2023self} & $74.6$ & $69.5$ & $60.0$ & $47.6$ & $31.8$ & $56.7$ & $52.25$ & $33.67$ & $8.40$ & $33.76$ \\
    Self-DETR + LTP (Ours) & $75.3$ & $69.4$ & $60.2$ & $48.5$ & $32.9$ & $57.3$ & $53.05$ & $35.69$ & $8.93$ & $35.09$ \\
    \hline
    Deformable-DETR~\cite{xizhou2021deformable_detr} & $71.5$ & $66.0$ & $55.5$ & $43.5$ & $29.6$ & $53.2$ & $50.05$ & $33.44$ & $9.81$ & $33.27$ \\
    Deformable-DETR + LTP (Ours) & $73.8$ & $69.4$ & $60.6$ & $48.0$ & $33.7$ & $57.1$ & $52.86$ & $36.27$ & $\boldsymbol{10.88}$ & $35.81$ \\
    \hline
    TadTR$^\dagger$~\cite{liu2021tadtr} & $74.8$ & $69.1$ & $60.1$ & $46.6$ & $32.8$ & $56.7$ & $49.91$ & $33.43$ & $8.77$ & $33.23$ \\
    TadTR$^\dagger$ + LTP (Ours) & $75.5$ & $69.7$ & $60.5$ & $47.3$ & $33.4$ & $57.3$ & $52.61$ & $36.15$ & $9.73$ & $35.46$ \\
    \hline\hline
    \rowcolor{gray!25}\multicolumn{11}{l}{\textit{\textbf{Encoder-only Methods}}} \\
    \hline\hline
    ActionFormer~\cite{zhang2022actionformer} & $82.1$ & $77.8$ & $71.0$ & $59.4$ & $43.9$ & $66.8$ & $53.50$ & $36.20$ & $8.20$ & $35.60$ \\
    ActionFormer + LTP (Ours) & $82.3$ & $78.3$ & $72.1$ & $59.8$ & $44.2$ & $67.3$ & $54.72$ & $37.88$ & $9.01$ & $36.72$ \\
    \hline
    TriDet$^\dagger$~\cite{shi2023tridet} & $83.6$ & $80.1$ & $72.9$ & $62.4$ & $47.4$ & $69.3$ & $54.06$ & $36.45$ & $7.90$ & $35.56$ \\
    TriDet$^\dagger$ + LTP (Ours) & $\boldsymbol{84.2}$ & $\boldsymbol{81.0}$ & $\boldsymbol{73.5}$ & $\boldsymbol{63.0}$ & $\boldsymbol{48.8}$ & $\boldsymbol{70.1}$ & $\boldsymbol{55.22}$ & $\boldsymbol{37.98}$ & $8.79$ & $\boldsymbol{37.03}$ \\
    \hline\hline
\end{tabular}
\caption{\textbf{Comparison results with the transformer-based models on THUMOS14 and ActivityNet-v1.3.} 
 `$\dagger$' indicates our reproduced versions.
}
\label{tab:main}
\end{table*}
\endgroup

\subsection{Implementation Details}
In this section, we briefly describe the implementation details.
We recommend to refer to the thorough elaboration for the details provided in the supplementary materials.
\noindent\textbf{Datasets.} 
Our experiments are conducted on three human action benchmarks: Kinetics-400~\cite{kay2017kinetics}, THUMOS14~\cite{jiang2014thumos14} and ActivityNet-v1.3~\cite{caba2015activitynet}.

\vspace{3pt}
\noindent\textbf{Feature Extractor.} 
We use the features of I3D~\cite{carreira2017i3d} or TSN~\cite{wang2016tsn} pre-trained on Kinetics.
The way of extracting follows the backbone model.

\vspace{3pt}
\noindent\textbf{DETR Architectures.} 
In this paper, we introduce two types of DETR baselines: 1) DAB-DETR~\cite{shilong2022dab_detr}, 2) Deformable-DETR~\cite{xizhou2021deformable_detr} in a temporal version.
These baselines do not have any additional module so that we can solely show the benefits of LTP.
The configure of DAB-DETR follows Self-DETR while that of Deformable-DETR is aligned with TadTR.
all other configurations follow the DETR baselines~\cite{liu2021tadtr, shi2022react, kim2023self} for fair comparison.

\begingroup
\setlength{\tabcolsep}{2.25pt} % Default value: 6pt
\renewcommand{\arraystretch}{1.0} % Default value: 1
\begin{table}[t]
\centering
\begin{tabular}{c|c|c||ccc|c||ccc|c}
    \hline\hline
    \multirow{2}{*}{B} & \multirow{2}{*}{O} & \multirow{2}{*}{S} &
    \multicolumn{4}{c||}{THUMOS14} & \multicolumn{4}{c}{ActivityNet-v1.3} \\
    \cline{4-11}
    & & & $0.3$ & $0.5$ & $0.7$ & Avg. & $0.5$ & $0.75$ & $0.95$ & Avg. \\
    \hline\hline
    $\cdot$ & $\cdot$ & $\cdot$ & $70.5$ & $53.9$ & $23.8$ & $50.3$ & $52.62$ & $32.59$ & $7.43$ & $32.40$ \\
    \checkmark & $\cdot$ & $\cdot$ & $71.5$ & $55.9$ & $29.9$ & $53.7$ & $52.29$ & $34.08$ & $8.41$ & $33.88$ \\
    \checkmark & \checkmark & $\cdot$ & $72.2$ & $57.8$ & $29.7$ & $54.2$ & $52.62$ & $34.99$ & $8.80$ & $34.61$ \\
    \checkmark & $\cdot$ & \checkmark & $73.6$ & $59.0$ & $33.7$ & $56.5$ & $52.83$ & $35.20$ & $9.03$ & $34.79$ \\
    \checkmark & \checkmark & \checkmark & $75.2$ & $60.4$ & $32.8$ & $57.3$ & $52.71$ & $35.60$ & $9.08$ & $34.92$ \\
    \hline\hline
\end{tabular}
\caption{\textbf{Ablation on pretext tasks.} 
`B', `O', `S' indicate Base, Ordinal, and Scale tasks, respectively.
}
\label{tab:ablation_condition}
\end{table}
\endgroup

\vspace{3pt}
\noindent\textbf{Pre-training Set-up.} 
For pre-training, we utilize the Kinetics-400 dataset~\cite{kay2017kinetics}, the same dataset on which the feature extractor is pre-trained, which contains approximately 300K videos.
We set the condition probability $p_{\text{cond}}$ for long-term pretext tasks as $0.5$..
We train the DETR model with a mini-batch size of 256 by using AdamW with initial learning rate of $1.0\times10^{-4}$ during 15 epochs.

\vspace{3pt}
\noindent\textbf{Fine-tuning Set-up.} 
As down-stream tasks, the pre-trained models are fine-tuned on ActivityNet and THUMOS14.
The configuration of fine-tuning follows the baselines.

\subsection{Main Results}

\noindent\textbf{Comparison with the State-of-the-Art.} 
We compare LTP applied on DAB-DETR~\cite{shilong2022dab_detr}, Deformable-DETR~\cite{xizhou2021deformable_detr} and Self-DETR~\cite{kim2023self} with the state-of-the art methods.
Table.~\ref{tab:main} demonstrates that our LTP significantly improves the performances of various DETR models on both benchmarks.

On THUMOS14, the DETR models with LTP have reached the state-of-the-art performance.
Especially for APs at high IoU thresholds such as $0.6$ and $0.7$, LTP remarkably improves the scores.
We conjecture that long-term pretext tasks in the pre-training enable the model to precise localization.
Interestingly for ActivityNet, the DETR-based models with LTP improve the performances with a large margin.
Similar to the results of THUMOS14, LTP has significantly improved the APs at high IoU thresholds especially at $0.95$.

Moreover, the table shows the results of the encoder-only models.
We found that they also suffer from the data scarcity problem.
As in the table, LTP significantly improves the performance by alleviating the issues from the data scarcity problem.

\begingroup
\setlength{\tabcolsep}{9.85pt} % Default value: 6pt
\renewcommand{\arraystretch}{1.0} % Default value: 1
\begin{table}[t]
\centering
\begin{tabular}{c|ccc|c}
    \hline\hline
    Synthesis & $0.50$ & $0.75$ & $0.95$ & Avg. \\ \hline\hline
    none & $52.62$ & $32.59$ & $7.43$ & $32.40$ \\
    instance-wise & $50.35$ & $32.54$ & $8.26$ & $32.62$ \\
    class-wise & $52.71$ & $35.60$ & $9.08$ & $34.92$ \\
    \hline\hline
\end{tabular}
\caption{\textbf{Alternative to synthesis.}
Experiments are conducted on ActivityNet-v1.3 with DAB-DETR.}
\label{tab:ablation_synthesis}
\end{table}
\endgroup

\noindent\textbf{Conditional Tasks.} 
We have conducted experiments on ActivityNet with DAB-DETR to prove the fidelity of each conditional task of LTP.
Table.~\ref{tab:ablation_condition} shows the results of the ablation study on pretext tasks of LTP on THUMOS14 and ActivityNet with DAB-DETR.
Each conditional task shows a clear improvement than only using the basic task.
This demonstrates that the conditional pretext tasks can help for learning long-term dependency allowing for more precise localization.
Also, the performance gain brought from the scale condition is slightly larger than the ordinal one.
We conjecture that the scale condition effectively eases the problem of imbalanced performances over scales.

\vspace{3pt}
\noindent\textbf{Synthesis.} 
In order to validate the benefits of class-wise synthesis in LTP, we further study on another alternative for synthesis.
One of the promising synthesis ways is the instance-wise method as in UP-DETR~\cite{dai2021up-detr}.
To this end, we randomly sample temporal intervals from videos.
Then the features of the interval are given as the condition, added to the action queries after projection in the same way as LTP.

Table.~\ref{tab:ablation_synthesis} shows the results for the alternative to class-wise synthesis on ActivityNet with DAB-DETR.
As shown, the instance-wise synthesis does not bring noticeable performance gain for the DETR model.
This is mainly because instance-wise synthesis for TAD handles a relatively short-term task than class-wise one so that it can suffer from a large task discrepancy between pre-training and fine-tuning.
Therefore, we claim that class-wise synthesis is the key point for pre-training of DETR for TAD.

\begingroup
\setlength{\tabcolsep}{10.85pt} % Default value: 6pt
\renewcommand{\arraystretch}{1.0} % Default value: 1
\begin{table}[t]
\centering
\begin{tabular}{l|ccc|l}
    \hline\hline
    Method & $0.50$ & $0.75$ & $0.95$ & Avg. \\ \hline\hline
    I3D & $52.62$ & $32.59$ & $7.43$ & $32.40$ \\
    I3D + LTP & $52.71$ & $35.60$ & $9.08$ & $34.92$ \\
    \hline
    TSP & $50.74$ & $32.89$ & $7.86$ & $32.83$ \\
    TSP + LTP & $53.80$ & $35.95$ & $8.45$ & $35.42$ \\
    \hline\hline
\end{tabular}
\caption{\textbf{Complementary to TSP.}
It shows LTP is complementary to TSP on ActivityNet-v1.3 with DAB-DETR.}
\label{tab:complementary}
\end{table}
\endgroup

\begingroup
\setlength{\tabcolsep}{1.45pt} % Default value: 6pt
\renewcommand{\arraystretch}{1.00} % Default value: 1
\begin{table}[t]
	\centering
	\begin{tabular}{l|ccccc|cccc}
		\hline\hline
        \multirow{2}{*}{Method} & \multicolumn{5}{c|}{Coverage} & \multicolumn{4}{c}{\# of Instances} \\
        \cline{2-10}
		& XS & S & M & L & XL & XS & S & M & L \\ \hline\hline
		BMN & $8.7$ & $24.4$ & $35.0$ & $55.2$ & $70.7$ & $53.4$ & $16.1$ & $9.8$ & $5.1$ \\
		GTAD & $8.4$ & $27.4$ & $39.0$ & $59.2$ & $71.9$ & $55.7$ & $17.7$ & $9.5$ & $2.9$ \\
		RCL & $11.8$ & $27.8$ & $39.4$ & $58.0$ & $71.1$ & $55.8$ & $19.4$ & $13.6$ & $5.1$ \\
        \hline
		DAB-DETR & $8.1$ & $15.9$ & $30.1$ & $56.4$ & $72.8$ & $53.4$ & $13.0$ & $9.6$ & $3.9$ \\
		+ LTP & $12.3$ & $20.0$ & $30.3$ & $55.4$ & $74.3$ & $54.3$ & $16.1$ & $15.7$ & $10.3$ \\
		\hline\hline
	\end{tabular}
	\caption{\textbf{Imbalanced performances.}
	We conduct DETAD analysis on ActivityNet-v1.3 based on DAB-DETR.}
	\label{tab:detad}
\end{table}
\endgroup

\begin{figure}[t]
\centering
\includegraphics[width=8.40cm]{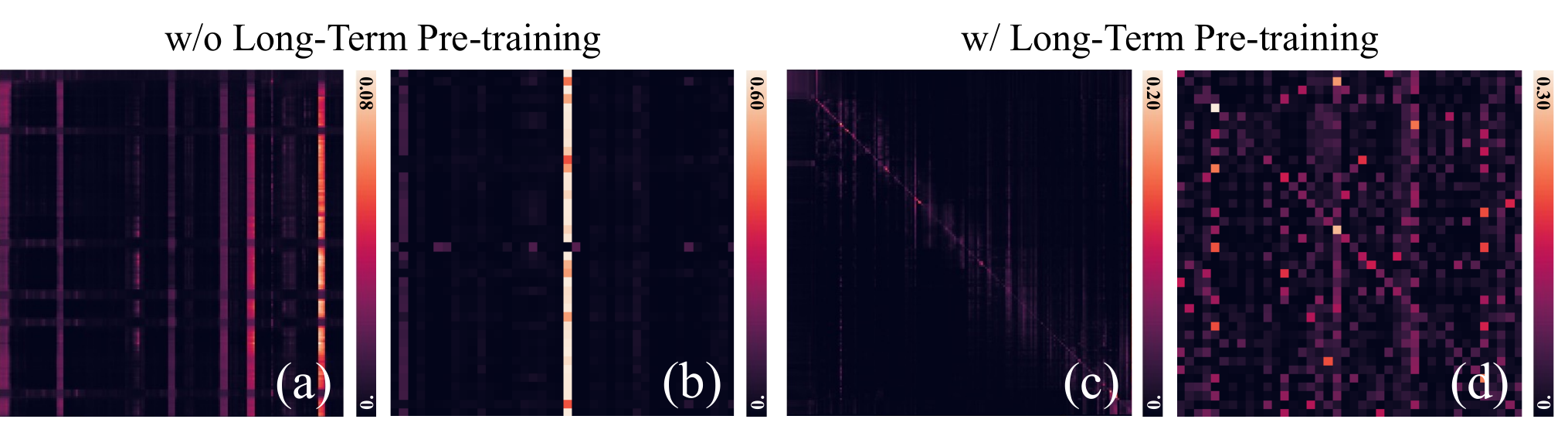}
\caption{\textbf{Attention maps.}
It shows self-attention maps from the last layers of the DAB-DETR encoder ((a), (c)) and decoder ((b), (d)) in test samples of ActivityNet-v1.3.
}
\label{fig:attention}
\end{figure}

\vspace{3pt}
\noindent\textbf{Complementary to TSP.} 
Table.~\ref{tab:complementary} shows the results of LTP with TSP~\cite{alwassel2021tsp}, a representative pre-training method for the feature extractor, on ActivityNet-v1.3 with DAB-DETR.
The table demonstrates that the performance improvement achieved by TSP persists in the integration of TSP and LTP, suggesting that the benefits of LTP complement pre-training the feature extractor.

\subsection{Analysis}

\vspace{3pt}
\noindent\textbf{Imbalanced Performances.} 
We observed that the model demonstrates significant imbalances in performance, primarily attributed to the limited data, particularly concerning the lengths of actions.
Table.~\ref{tab:detad} shows the sensitivity analysis of DETAD~\cite{alwassel2018detad} in terms of coverage (relative scale to video) and the number of instances.
It illustrates the imbalanced performances of DAB-DETR, indicating a higher likelihood of the model predicting long actions. 
As a result, the performances for smaller scales and videos with numerous short instances are notably low.

\begin{figure}[t]
\centering
\includegraphics[width=8.35cm]{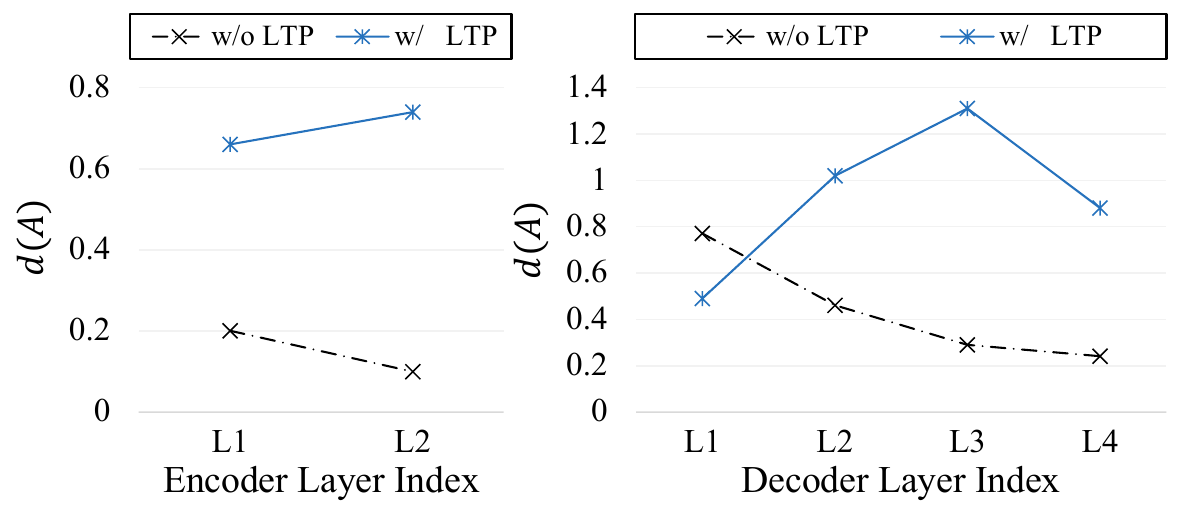}
\caption{\textbf{Diversity of self-attention maps.}
To analyze the effect of our pre-training for the attention collapse, we measure the diversity defined in Eq.~\ref{eq:diversity} of the self-attention maps.
}
\label{fig:diversity}
\end{figure}

However, the integration of LTP into DAB-DETR significantly alleviates the imbalanced performances, particularly in extra-short (XS) and short (S) scales.
Of greater interest, the performances for videos with medium (M) and large (L) numbers of action instances exhibited remarkable improvement, achieving a new state-of-the-art score.
This is primarily attributed to LTP preventing the model from overfitting to the action distribution of ActivityNet, which predominantly consists of a single long instance in the video.

\vspace{3pt}
\noindent\textbf{Attention Collapse.} 
Fig.~\ref{fig:attention} shows self-attention maps from the last layers of the DAB-DETR encoder ((a), (c)), and decoder ((b), (d)) for test samples of ActivityNet-v1.3.
We observe that DETR for TAD severely suffers from the attention collapse problem.
As in the attention maps ((a), (b)) of Fig.~\ref{fig:attention}, all queries of DETR without LTP attend a few key elements.
However, the attention maps ((c), (d)) of the model with LTP do not exhibit collapsed relations; instead, tokens show correlations with themselves or neighboring tokens.

Furthermore, we quantify the diversity of self-attention maps according to \cite{dong2021rank_collapse, kim2023self}.
The diversity $d(A)$ for the attention map $A$ is the measure of the closeness between the attention map and a rank-1 matrix as defined as below:
\begin{equation}
    \begin{aligned}
    d(A) = \| A - \boldsymbol{1}a^\top \|, \text{where}~a = \argmin_{a^{\prime}}\| A - \boldsymbol{1}a^{\prime\top} \|,
    \label{eq:diversity}
    \end{aligned}
\end{equation}
where $\| \cdot \|$ denotes the $\ell_1$,$\ell_\infty$-composite matrix norm, $a$, $a'$ are column vectors of the attention map $A$, and $\boldsymbol{1}$ is an all-ones vector.
Note that the rank of $\boldsymbol{1}a^\top$ is 1, and therefore, a smaller value of $d(A)$ means $A$ is closer to a rank-1 matrix.

Fig.~\ref{fig:diversity} shows the diversity on each layer of the encoder and decoder for DAB-DETR models with and without LTP.
The diversity is measured by the average value over all videos in the test set of ActivityNet-v1.3.
As the model depth gets deeper, the diversity of the baseline decreases close to $0$.
However, the diversity of LTP keeps high, signifying that LTP eases the attention collapse.
While our pre-training method does not employ a direct solution to the collapse, such as Self-DETR~\cite{kim2023self}, the results suggest that furnishing the model with sufficient data diversity can alleviate the problem and lead to a more substantial performance improvement described in Table.~\ref{tab:main}.

%% file: sec/5_conclusion.tex
\section{Conclusion}
In this paper, we have proposed a new pre-training strategy, Long-Term Pre-training (LTP) tailored for DETR in TAD, to migrate the problems caused by data scarcity.
LTP has two main components: 1) class-wise synthesis, 2) long-term pretext tasks.
Class-wise synthesis is to construct video features based on a target category to minimize the task discrepancy.
In addition, two conditional pretext tasks foster the DETR model to learn long-term dependency.
The extensive experiments have demonstrated that LTP relieves the problems of data scarcity with the remarkably improved performances.

%% file: sec/7_supplementary.tex
\renewcommand{\thepage}{A\arabic{page}}  
\renewcommand{\thesection}{A}   
\renewcommand{\thetable}{A\arabic{table}}   
\renewcommand{\thefigure}{A\arabic{figure}}

\setcounter{section}{0}
\setcounter{table}{0}
\setcounter{figure}{0}

\section{Additional Details}
\noindent\textbf{Feature Extractor.} 
We use the features of I3D or TSN pre-trained on Kinetics.
We follow video pre-processing process as in I3D, which uses 25 FPS for frame extraction and resizes the frames so that they have 256 pixels in their longer side (width or height).
We extract temporal features with 16-frame non-overlapping windows, and use the final outputs with global average pooling (GAP).
Each temporal feature is corresponding to 16 frames since the final temporal resolution of I3D is $1/16$ of the input video.
Note that the feature extractor is fixed during pre-training and fine-tuning.
The way of extracting follows the backbone model.

\vspace{3pt}
\noindent\textbf{DETR Architectures.} 
In this paper, we introduce two types of DETR baselines: 1) DAB-DETR, 2) Deformable-DETR in a temporal version.
These baselines do not have any additional module so that we can solely show the benefits of LTP.
The configure of DAB-DETR follows Self-DETR while that of Deformable-DETR is aligned with TadTR.
The hidden dimension of DETR models is $256$ as the DETR baselines for TAD.
The number of queries is $40$.
Also, the number of the encoder and decoder layers is $2$, and $4$, respectively.
all other configurations follow the DETR baselines~\cite{liu2021tadtr, shi2022react, kim2023self} for fair comparison.

\vspace{3pt}
\noindent\textbf{Pre-training Set-up.} 
For pre-training, we utilize the Kinetics-400 dataset, the same dataset on which the feature extractor is pre-trained, which contains approximately 300K videos.
The length of the temporal features to synthesize is $192$, which is the same as that of ActivityNet-v1.3.
Also, the maximum number of action instances $N_{\text{max}}$ is $12$ since the temporal length of each video is $16$.
We set the condition probability $p_{\text{cond}}$ for long-term pretext tasks as $0.5$.
When two videos of the same target class are overlapped in class-wise synthesis, we group them as a single long action instance.
We train the DETR model with a mini-batch size of 256 by using AdamW with initial learning rate of $1.0\times10^{-4}$ during 15 epochs.
The learning rate is scheduled by cosine annealing with a warm-up of 5 epochs.

\vspace{3pt}
\noindent\textbf{Fine-tuning Set-up.} 
As down-stream tasks, the pre-trained models are fine-tuned on ActivityNet and THUMOS14.
The configuration of fine-tuning follows the baselines.
Note that the parameters of the classification head are not initialized with the pre-train weights because the number of classes does not match.
Also, we do not add any vector to the action queries as condition when fine-tuning.
The lengths of the temporal features are $192$, and $128$ for ActivityNet-v1.3, and THUMOS14, respectively.
We fine-tune the DETR model with a mini-batch size of 16 by using the AdamW optimizer for both benchmarks.
As for ActivityNet, we train the model with the learning rate of $1.0\times10^{-4}$ during 20 epochs.
Also, we schedule the learning rate by cosine annealing with a warm-up of 5 epochs.
On THUMOS14, we train the model with the learning rate of $2.0\times10^{-4}$ during 120 epochs.
We schedule the learning rate multiplied by 0.1 at 80 and 100 epochs, respectively.

\vspace{3pt}
\noindent\textbf{Details for the Synthesis Ablation.} 
In Table.~2 of the paper, we have conducted experiments for an alternative to the class-wise synthesis of LTP.
We employ the way of UP-DETR as the alternative synthesis, named `instance-wise' in the table.
To be more specific, we first randomly sample features from a video in the dataset.
Then we randomly select an interval to temporally crop the features as the target.
The cropped features are average-pooled and added to the action queries as a condition.
The DETR model is trained to find the interval from the entire video as UP-DETR.
As shown in Table.~2 of the paper, we found that this way is not valid for TAD.

\begingroup
\setlength{\tabcolsep}{2.40pt} % Default value: 6pt
\renewcommand{\arraystretch}{1.00} % Default value: 1
\begin{table}[t]
	\centering
	\begin{tabular}{c|ccccc|cccc}
		\hline\hline
        \multirow{2}{*}{Method} & \multicolumn{5}{c|}{Coverage} & \multicolumn{4}{c}{\# of Instances} \\
        \cline{2-10}
		& XS & S & M & L & XL & XS & S & M & L \\ \hline\hline
		BMN & $8.7$ & $24.4$ & $35.0$ & $55.2$ & $70.7$ & $53.4$ & $16.1$ & $9.8$ & $5.1$ \\
		GTAD & $8.4$ & $27.4$ & $39.0$ & $59.2$ & $71.9$ & $55.7$ & $17.7$ & $9.5$ & $2.9$ \\
		RCL & $11.8$ & $27.8$ & $39.4$ & $58.0$ & $71.1$ & $55.8$ & $19.4$ & $13.6$ & $5.1$ \\
        \hline
		w/o LTP & $12.1$ & $19.4$ & $29.2$ & $52.3$ & $70.7$ & $51.7$ & $15.6$ & $15.2$ & $8.8$ \\
		w/~~~LTP & $13.2$ & $22.9$ & $33.8$ & $56.3$ & $74.7$ & $55.2$ & $18.3$ & $16.3$ & $9.4$ \\
		\hline\hline
	\end{tabular}
	\caption{\textbf{DETAD analysis.}
	We conduct DETAD analysis on ActivityNet-v1.3 based on Deformable-DETR.}
	\label{tab:detad_deform}
\end{table}
\endgroup

\begin{figure*}[t]
\centering
\includegraphics[width=16.35cm]{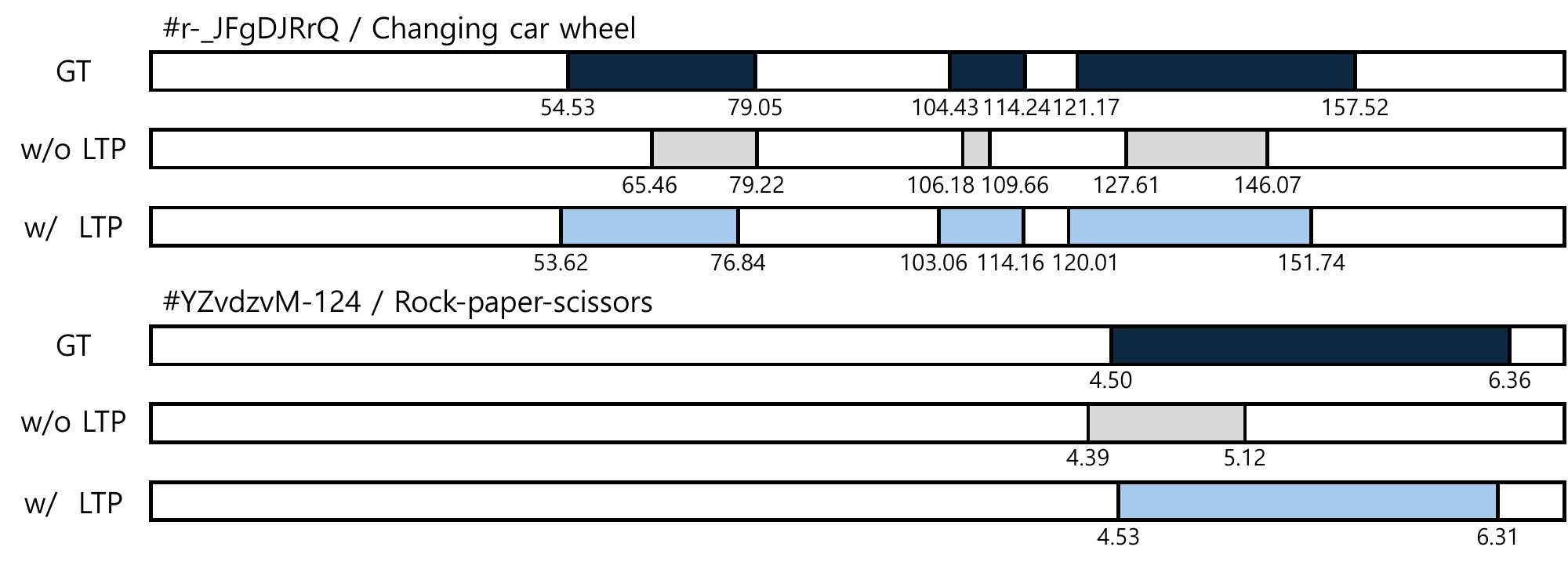}
\caption{\textbf{Qualitative results with and without LTP. }
The figure depicts test samples on ActivityNet-v1.3 with DAB-DETR.
}
\label{fig:qualitative_results}
\end{figure*}

\vspace{3pt}
\noindent\textbf{Details for the Experiments with TSP.} 
In Table.~5 of the paper, we have validated the complementary benefits with TSP~\cite{alwassel2021tsp}, a pre-training method for the feature extractor.
To this end, we first extract TSP features of the Kinetics-400 dataset.
We use the official codes from \cite{alwassel2021tsp} with a stride 16, which means a non-overlapping set-up.
After feature extraction, we pre-train and fine-tune the DAB-DETR model as done in the paper.

\section{Additional Results}

\noindent\textbf{DETAD Analysis for Deformable-DETR.} 
Table.~\ref{tab:detad_deform} shows the results of the DETAD~\cite{alwassel2018detad} analysis on ActivityNet-v1.3 with Deformable-DETR.
Similar to the results of DAB-DETR in Table.~4 of the paper, LTP significantly improves the performances of extra-small (XS) and small (S) actions.
Moreover, the performances for the middle (M) and large (L) numbers of actions instances are largely boosted by LTP.

\begin{figure}[t]
\centering
\includegraphics[width=8.35cm]{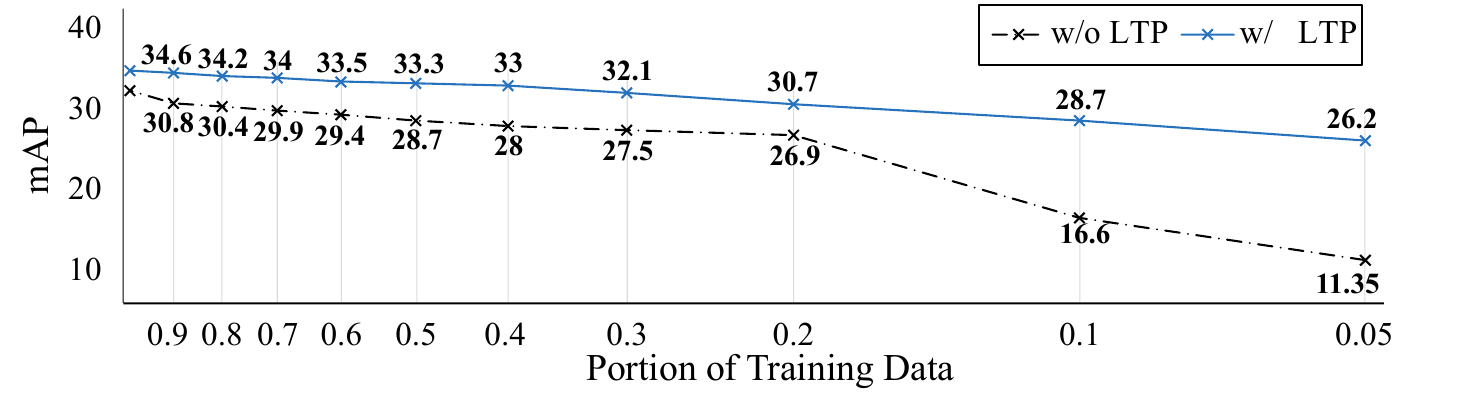}
\caption{\textbf{Data scarcity.}
We conducted experiments for training data ablation of ActivityNet-v1.3 with DAB-DETR.
}
\label{fig:data_scarcity}
\end{figure}

\vspace{3pt}
\noindent\textbf{Data Scarcity. }
To further analyze the benefits of LTP under data scarcity, we conducted additional experiments using a reduced amount of training data from ActivityNet-v1.3 with DAB-DETR.
To achieve this, we randomly sampled and utilized training data ranging from 5\% to 90\%, while maintaining the entirety of the test set.

Fig.~\ref{fig:data_scarcity} presents the results in mAP based on the DAB-DETR model, both with and without LTP.
DETR with LTP consistently outperforms the baseline without LTP in terms of mAP for all configurations.
Despite the rapid degradation in the performance of DAB-DETR without LTP, the model with LTP exhibits a much slower decrease in performance.
This demonstrates that long-term pre-training effectively mitigates the issue of data scarcity.

\vspace{3pt}
\noindent\textbf{Qualitative Results. }
Fig.~\ref{fig:qualitative_results} illustrates test samples in ActivityNet-v1.3 based on DAB-DETR models with and without LTP.
As shown, DAB-DETR with LTP outputs more precise results.

\begin{figure}[t]
\centering
\includegraphics[width=8.40cm]{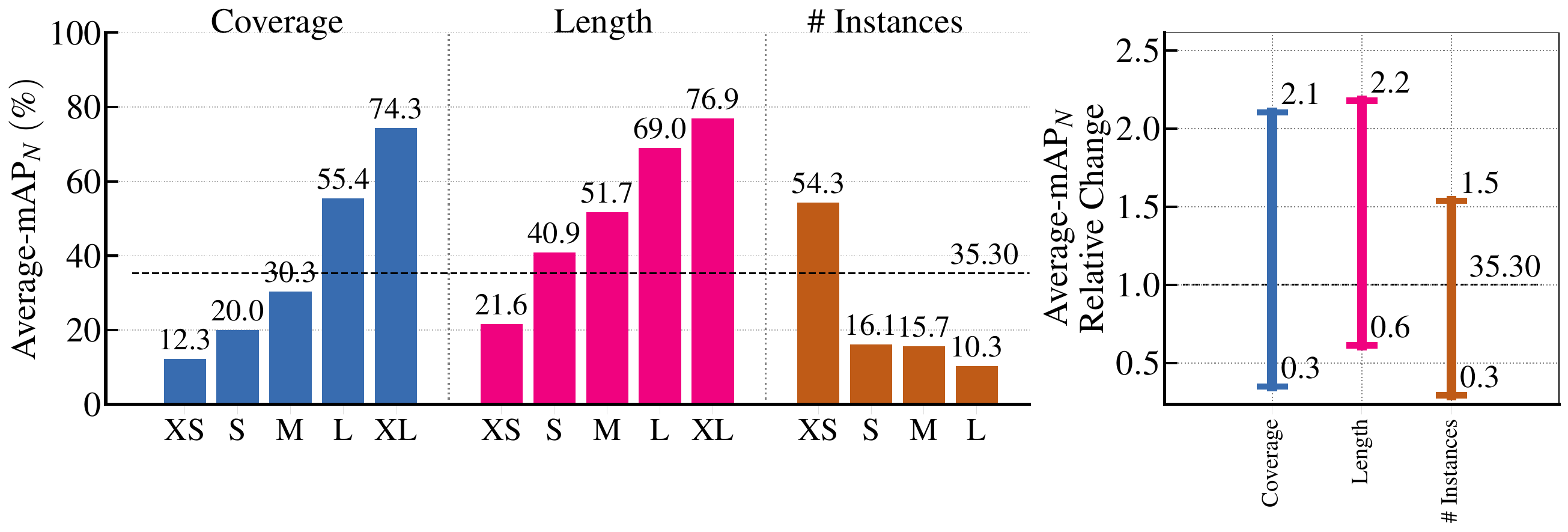}
\caption{\textbf{DETAD analysis on ActivityNet.} It shows the DETAD~\cite{alwassel2018detad} sensitivity analysis on ActivityNet with DAB-DETR.
}
\label{fig:detad_sensitivity_analysis_DAB}
\end{figure}

\vspace{3pt}
\noindent\textbf{Visualization of DETAD Analysis for DAB-DETR.} 
Fig.~\ref{fig:detad_sensitivity_analysis_DAB} depicts the results of DETAD~\cite{alwassel2018detad} sensitivity analysis for DAB-DETR on ActivityNet-v1.3.